\documentclass[preprint,12pt]{elsarticle}

\usepackage{amssymb,url,natbib,graphicx,subfigure,url}

\newtheorem{prop}{Proposition}
\journal{***}

\begin{document}

\begin{frontmatter}



\title{Total Variation, Adaptive Total Variation and Nonconvex Smoothly Clipped Absolute Deviation Penalty for Denoising Blocky Images }

\author[label1]{Aditya Chopra\fnref{label3}}
\author[label2]{Heng Lian}
\address[label1]{School of Computing Sciences, VIT University, Vellore, TN, India}
\address[label2]{Division of Mathematical Sciences, School of Physical and Mathematical Sciences, Nanyang Technological University, Singapore, 637371}

\author{}

\address{}

\begin{abstract}
The total variation-based image denoising model has been generalized and extended in numerous ways, improving its performance in different contexts. We propose a new penalty function motivated by the recent progress in the statistical literature on high-dimensional variable selection. Using a particular instantiation of the majorization-minimization algorithm, the optimization problem can be efficiently solved and the computational procedure realized is similar to the spatially adaptive total variation model. Our two-pixel image model shows theoretically that the new penalty function solves the bias problem inherent in the total variation model. The superior performance of the new penalty is demonstrated through several experiments. Our investigation is limited to ``blocky" images which have small total variation.
\end{abstract}

\begin{keyword}


MM algorithm\sep SCAD penalty\sep Total variation denoising. 
\end{keyword}

\end{frontmatter}


\section{Introduction}
\label{}
Denoising is probably the most common and most studied problem in image processing. Approaches developed so far include many methods arising from the field of engineering, computer science, statistics and applied mathematics. There are several popular classes of existing denoising algorithms, from simple linear neighborhood filtering to complicated wavelet method based on solid statistical foundation \citep{donoho95a,figueiredo01,portilla03}. PDE-based method proposed first in \cite{rudin91} is unique in its formulation of images as functions in a suitable function space. Relatively few comparison studies exist among different methods, which is quite understandable due to (1) there are a large number of existing denoising approach with many different modifications and extensions; (2) the success or failure of different approaches depend largely on the characteristics exhibited by different types of images, whether cartoon or natural scene images, grayscale or colored, textured or solid objects. One exception is the work \cite{nowak06} which compared the standard total variation (TV) model with wavelet denoising and find TV is inferior for some standard test images. With different fine tuning and extensions available in both the class of PDE-based and  wavelet-based methods, such as using higher order derivatives or correlated wavelet coefficients, it is still hard to judge from their results the relative merits of these two approaches, although it seems to be the prevailing mindset that the wavelet-based methods works better for general images. 

Denoting the unobserved original noiseless image by $u$, the goal of denoising is to recover this original image given an observed noisy image $f=u+n$, where $n$ denotes the noise. In traditional filtering as well as wavelet-based approaches, we either think of images as $m\times n$ matrices or $N=mn$-dimensional vectors, while the PDE-based method will generally treat images as bivariate functions defined on the unit square $\Omega=[0,1]\times [0,1]$. Introduced in \cite{rudin91}, the standard total variation (TV) image denoising method estimates the original image by solving the following minimization problem
\begin{equation}\label{eqn:tv}
\hat{u}=\arg\min_u ||f-u||^2+\lambda TV(u),
\end{equation}
where $||.||^2$ is the $L_2$ norm of the function and $TV(u)=\int_\Omega|\nabla u|$ is the total variation norm of $u$ \cite{rudin91}. The regularization parameter $\lambda$ controls the tradeoff between the fidelity to observed image and smoothness of the recovered image. Actually the paper \cite{rudin91} used the somewhat equivalent formulation of minimizing the total variation with constraints on the noise level, which is assumed to be known. But the penalized $L_2$ version stated above is more convenient when the level of the noise is unknown and we will adopt this formulation in our study. Both practically and theoretically, this model is the best understood one in PDE-based methods as of today, where the images are considered as belonging to the space of functions of bounded variation (BV) and the existence and uniqueness of solution is well-established \cite{lions97,vese01,dobson96}. Discrete version of the TV model is considered in \cite{nowak06}, arguing that all approaches have to go through the discretization procedure when implemented anyway. Our point of view is that using either the continuous or discrete formulation for the PDE-based method makes little difference in practice.

Although the standard TV model above might not be competitive for general image denoising tasks, it is believed to be ideal for blocky images, i.e., images that are nearly piece-wise constant. From a statistical point of view, this can be simply seen by the fact that it penalizes the first partial derivative (or, in discrete version, first order differences) and thus shrinks them to zero. \cite{strong97} noted the inherent bias in TV model and proposed the spatially adaptive total variation (SATV) model that applies less smoothing near significant edges by utilizing a spatially varying weight function that is inversely proportional to the magnitude of image derivatives. SATV is a two-step procedure where the weight function obtained from the first step using standard TV is then used to guide smoothing in the second step. The authors showed that with a modest increase in computation, SATV is superior to standard TV in restoring piece-wise constant image features. 

Curiously, there is an almost parallel development in the statistical literature in the context of high-dimensional linear regression with variable selection. As explained in the next section, these studies focus on the regression problem where although there exists a priori numerous covariates, most of the regression coefficients are exactly zero, implying that the corresponding covariates have no effects on the response variable. Thus shrinking most regression coefficients to zero is a viable strategy for efficient estimation. For piece-wise constant images, with first derivatives in most locations exactly equal to zero, shrinking them to zero is thus also a reasonable approach. Taking advantage of this observation, we propose to adapt the smoothly clipped absolute deviation (SCAD) penalty \cite{fan01,fan04} that has become extremely popular in the statistical community for our image denoising task. Although in the case of TV model the correspondence between the functional-analytical approach and the statistical approach seems to be well-known, and some have studied in detail the properties of total variation from a statistical point of view \cite{mammen97,davies01}, these statistical works are only restricted to the one-dimensional case. Besides, as far as we know the parallelism stated above has not been fully utilized and in particular the SCAD penalty has not been applied to penalize the first order differences even in the one-dimensional case. Besides its superior performance in practice, there are several advantages of SCAD penalty compared to SATV, most notably getting rid of the extra parameter that a user needs to tune for SATV in implementation. As mentioned before, we think either discrete or continuous formulation formally makes little difference, but we choose to use the continuous formulation since it can simplify description and notation significantly. The only problem is that the functional using SCAD penalty being nonconvex, existence of solution is not guaranteed. The theoretically inclined reader might want to think in discrete terms so that such technical point does not arise. Our computational experiments show that SCAD is superior to SATV in terms of mean square error (MSE). Although MSE is notorious for describing the visual quality of an image, it is arguably less so for blocky images where MSE can describe the accuracy of restoration rather faithfully. 

The rest of the paper is organized as follows. In the next section, we briefly review the TV and the SATV model and point out the almost trivial connection to Lasso and the adaptive Lasso developed in the statistical literature so that we hope readers from both fields can follow the motivation and  development of the current paper. In Section 3, we adapt the SCAD penalty for our image denoising problem and discuss some properties in detail in this context. We also developed a majorization-minimization procedure using first order Taylor expansion so that the computation involved simply reduces to that similar to the SATV model, although with a different weight function.   In Section 4, we will briefly review a method called Monte-Carlo SURE \cite{ramani08} for regularization parameter selection which is used in our study when required. In Section 5, several computational experiments are used to show the superiority of the proposed method in denoising blocky images. In these experiments, we also intentionally emphasize the difficulty encountered with SATV model in tuning its performance. We conclude the paper with a discussion in Section 6.

\section{Review of the TV and SATV model}
The TV model proposed by \cite{rudin91} and presented above in equation (\ref{eqn:tv}) has received a great deal of attention in the last decade. In \cite{strong97}, the authors argued that it is desirable that less smoothing is carried out where there is more feature in the image. This motivated the replacement of TV norm by the following more general weighted TV functional
\begin{equation}\label{eqn:satv}
TV_w(u)=\int_\Omega w(x,y)|\nabla u(x,y)|\,dxdy.
\end{equation}
The weight should be small in the presence of an edge so that less smoothing is performed near an edge. \cite{strong97} used a weight function inversely proportional to the derivative, with a parameter $e$ added both to avoid dividing by zero and to be used as a tuning parameter to control the amount of adaptivity. Thus in their proposal of the spatially adaptive total variation (SATV) model $w=1/(u_x+e)+1/(u_y+e)$ where $u_x$ and $u_y$ are the partial derivatives. \cite{strong97} used a two-step method. In the first step the standard TV model (\ref{eqn:tv}) is used to estimate $u$ based on which the partial derivatives (first order differences) are computed. Then the derivatives are used in (\ref{eqn:satv}) to compute the final restored image. If $e$ is chosen sufficiently large, SATV basically reduces to the standard TV. On the other hand, if $e$ is too small, artificial edges will appear and the algorithm will be numerically unstable as well. We will see in our simulations that the result is somewhat sensitive to the choice of $e$ and the appropriate amount of adaptivity is not universal to all images, which makes it difficult to choose $e$ in practice, or leads to a sizable increase of the amount of computation required to say the least. 

As we mentioned in the introduction, there is an almost parallel line of development in the statistical literature that uses the same idea of SATV in a different context. In a linear regression problem ${y}_i=\mathbf{x}_i^T\beta+\epsilon_i$ based on independent and identically  distributed (i.i.d.) data $({y}_i,\mathbf{x}_i)_{i=1}^n$, where $\mathbf{x}_i=(x_{i1},\ldots,x_{ip})^T$ are the covariates, $\beta=(\beta_1,\ldots,\beta_p)^T$ are the regression coefficients, and $\epsilon_i$ is a zero mean noise. Sometimes one has good reasons to believe that only a few of the $x_{iq}$'s are related to $y_i$, i.e., many of the $\beta_q$'s are exactly zero. In these situations it is desirable to design an approach that shrinks many regression coefficients to zero automatically. Lasso \citep{tibshirani96} does exactly that and is formulated as the minimization of the following objective function:
\[\sum_{i=1}^n||y_i-\mathbf{x}_i^T\beta||^2+\lambda \sum_{i=1}^p|\beta_i|. \]
It is now well-known that this algorithm encourages many coefficients to be exactly zero as desired due to the use of $L_1$ norm penalty for $\beta$. \cite{zou06} later proposes the adaptive Lasso, which possesses better theoretical properties than Lasso and also proves to be superior in practice, that solves the following minimization problem
\[\sum_{i=1}^n||y_i-\mathbf{x_i}^T\beta||^2+\lambda \sum_{i=1}^p|\beta_i|/|\hat{\beta}_i|, \]
where $\hat{\beta}=\{\hat{\beta_1},\ldots,\hat{\beta_p}\}$ is the standard least square estimate. Any other reasonable estimate can be used (to be more rigorous, $\hat{\beta}$ must be \textit{consistent} in statistical terms in order to enjoy the theoretical properties stated in that paper). 

The reader can immediately see the parallel developments in statistics and TV-based image processing. When it is desirable to shrink the first order differences in an image towards zero, the same arguments that lead to Lasso and adaptive Lasso now assume the form of TV and SATV respectively. In the statistical literature, \cite{mammen97,davies01} studied the TV problem in its discrete form, but we have not seen any mention of utilizing adaptive Lasso to penalize the first order differences. 

Historically, before the appearance of adaptive Lasso, to address the shortcomings
of Lasso (which is not consistent in variable selection), \cite{fan01} proposed the smoothly clipped absolute deviation (SCAD) penalty which is motivated by the desire to achieve several desirable properties of the estimator such as continuity, asymptotic unbiasedness, etc. They also show
that the resulting estimator possesses the so-called oracle property, i.e. it is consistent
for variable selection and behaves the same as when the zero coefficients are
known in advance. In the next section, we adapt the SCAD penalty for image processing tasks. Using SCAD penalty gets rid of the clumsiness of having to choose the parameter $e$ in SATV and our experiments show its performance is superior to SATV.

\section{Image Denoising with the SCAD penalty}
In linear regression, using the SCAD penalty amounts to minimizing the following functional
\begin{equation}\label{eqn:scadlsq}
\sum_{i=1}^n ||y_i-\mathbf{x}_i^T\beta||^2+\sum_{i=1}^p p_\lambda(|\beta_i|),
\end{equation}
where $p_\lambda(.)$ is more conveniently defined by its derivative
\[
p_\lambda'(\theta)=\lambda\left\{I(\theta\le\lambda)+\frac{(a\lambda-\theta)_+}{(a-1)\lambda}I(\theta>\lambda)\right\}, \mbox{ for } \theta>0, 
\]
 and $p_\lambda(0)=0$. As usual, $a=3.7$ is used.

\begin{figure}
\centerline{\subfigure[]{\includegraphics[width=2.5in]{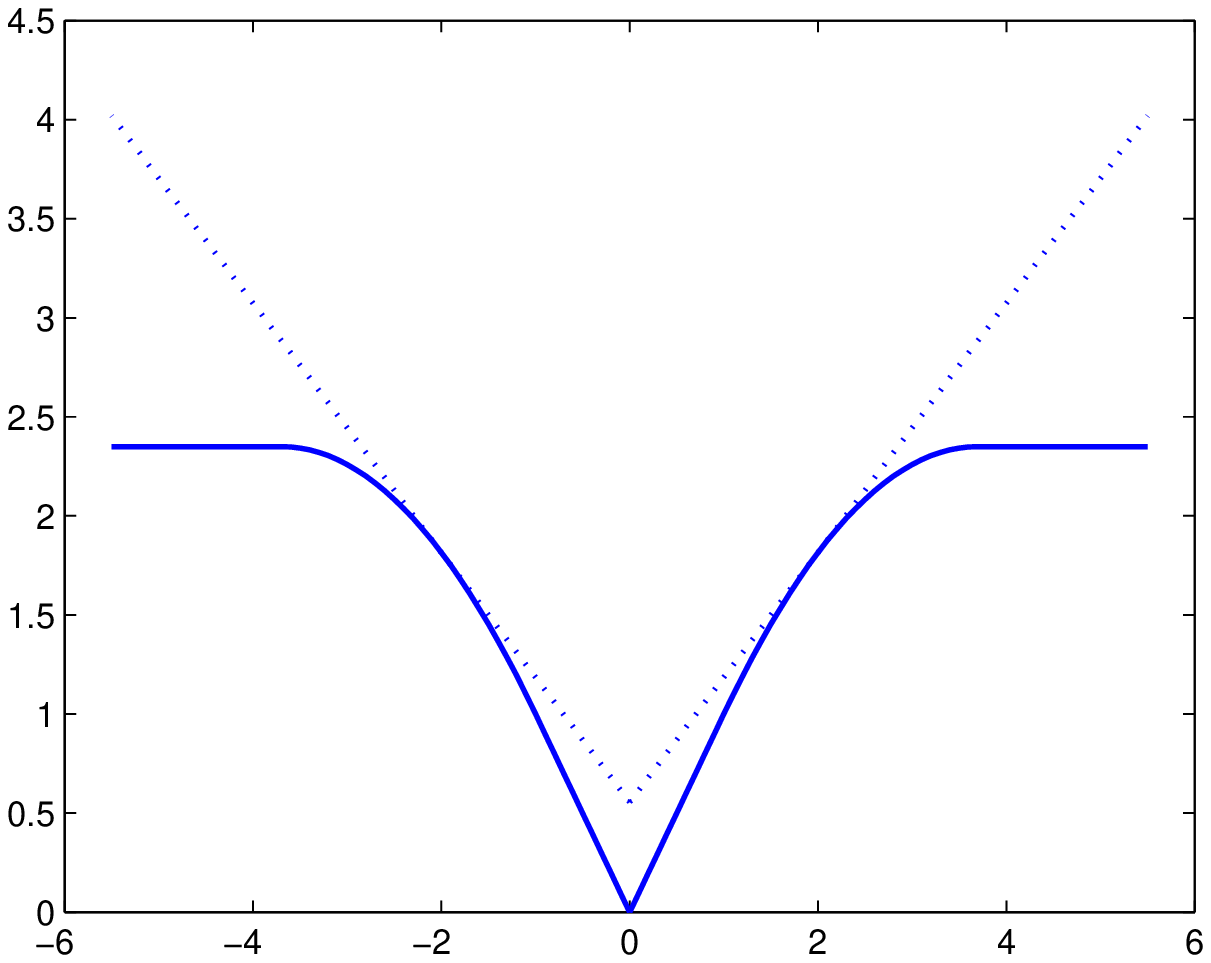}
}
\hfil
\subfigure[]{\includegraphics[width=2.5in]{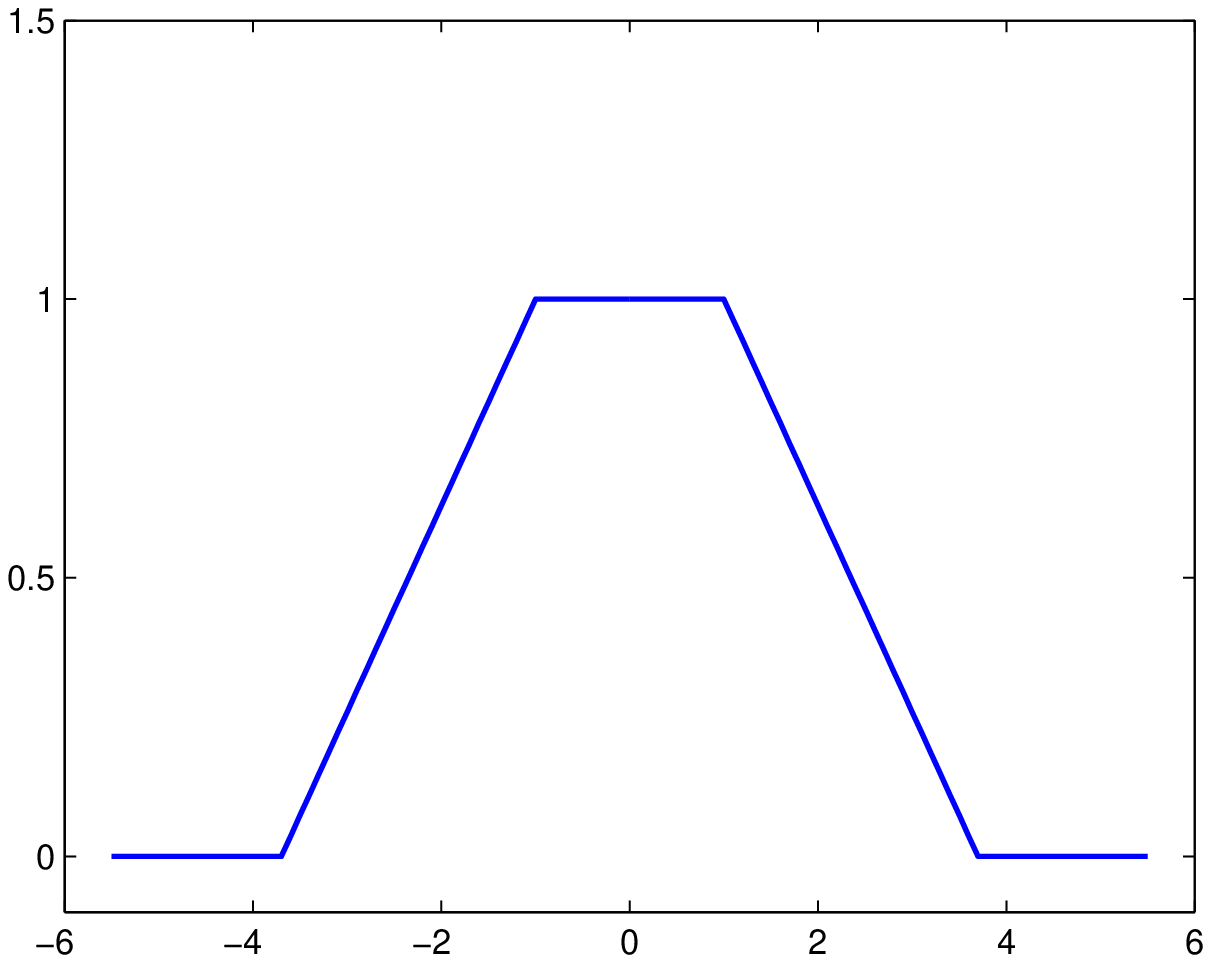}
}}
\caption{(a) The SCAD penalty function and its linear approximation. (b) The derivative of the SCAD penalty function.\label{scad}   }
\end{figure}

We plot the function $p_\lambda$ in Fig \ref{scad}(a) for $\lambda=1$ and its derivatives in Fig \ref{scad}(b). As seen in (\ref{eqn:scadlsq}) we only use $p_\lambda$ and its derivative with a nonnegative functional argument. We plot both in Fig \ref{scad} as even functions for convenience, although the derivative should be an odd function if $p_\lambda$ is defined as an even function. Note that this penalty function, unlike the $L_1$ penalty used in Lasso, is not convex.
To use the SCAD penalty for image denoising, we formally write down the functional
\begin{equation}\label{eqn:scad}
||f-u||^2+\int_\Omega p_\lambda(|\nabla u|).
\end{equation}
Some readers will have the objection that $p_\lambda$ is nonconvex and thus the existence of solution to the above functional is in question. Even the definition of $p_\lambda(|\nabla u|)$ seems to be a difficult task, if not impossible. Note \cite{vese01} only defined $\phi(|\nabla u|)$ when $\phi$ is convex and $u$ is a BV function. Due to this problem we encourage the reader to change to a discrete formulation which is straightforward from (\ref{eqn:scad}). The expression (\ref{eqn:scad}) in the continuous form is so much cleaner so we prefer to keep it. This should hopefully be just a minor nuisance for practitioners.

To see clearly the effect of SCAD compared to TV, we consider the following simple discrete problem instead,
\begin{equation}\label{eqn:twopixel}
\arg\min_{\theta_1,\theta_2}\;(y_1-\theta_1)^2+(y_2-\theta_2)^2+p_\lambda(|\theta_1-\theta_2|),
\end{equation}
i.e., we consider an ``image" with only two pixels. We have the following property of the minimizer comparing SCAD penalty and TV penalty, the proof is deferred to the appendix:
\begin{prop} Suppose without loss of generality that $y_1\ge y_2$. 

(a) If $y_1-y_2>a\lambda$, the minimizer of (\ref{eqn:twopixel}) is $\theta_1=y_1,\theta_2=y_2$.

(b) If $y_1-y_2<\min_{\xi\in R}(|\xi|+p_{\lambda}'(|\xi|)$, the minimizer of (\ref{eqn:twopixel}) is $\theta_1=\theta_2=(y_1+y_2)/2$. 

If instead the TV norm is used, i.e. $p_\lambda(|\theta_1-\theta_2|)$ is replaced by $\lambda|\theta_1-\theta_2|$ in (\ref{eqn:twopixel}), then

(c) if $y_1-y_2>\lambda$, the minimizer is $\theta_1=y_1-\lambda/2, \theta_2=y_2+\lambda/2$.

(d) if $y_1-y_2\le\lambda$, the minimizer is $\theta_1=\theta_2=(y_1+y_2)/2$.
\end{prop}

From the proposition, we see that for this simple two-pixel image model, although both penalties have the effect of shrinking $\theta_1$ and $\theta_2$ to be exactly equal to each other, the SCAD penalty has the additional desired property that when the difference $|y_1-y_2|$ is large enough, no shrinkage is applied. From part (c) of the proposition the TV model is implicitly biased, which is already known in more general contexts as shown in \cite{strong03,chan05}. Our experiments later also demonstrated this effect. From the proof in the Appendix it can be seen that this difference arises basically from the fact that $p_\lambda'(\theta)=0$ when $\theta$ is big enough.

Compared to TV or SATV, optimization of the functional (\ref{eqn:scad}) is more complicated since the functional is nonconvex and using time evolution of the corresponding Euler-Lagrange equation (i.e., gradient descent) is potentially problematic. Thus we use the following majorization-minimization (MM) algorithm instead. Note that \cite{nowak06} also proposed an MM algorithm for standard TV image denoising.

First, we majorize the SCAD penalty function using its first order Taylor expansion using an initial estimated image $u^{(0)}$ (we could simply set $u^{(0)}=f$ for example): 
\[p_\lambda(|\nabla u|)\le p_\lambda(|\nabla u^{(0)}|)+p_\lambda'(|\nabla u|)(|\nabla u|-|\nabla u^{(0)}|),\]
which is illustrated in Fig \ref{scad}(a) as the dotted line. Using this approximation, we can repeatedly solve the problem:
\[u^{(k)}=\arg\min_u ||f-u||^2+\int p_\lambda(|\nabla u^{(k-1)}|)+p_\lambda'(|\nabla u^{(k-1)}|)(|\nabla u|-|\nabla u^{(k-1)}|), k=1,2,\ldots,K,\]
i.e., replacing the SCAD penalty by its upper bound and then solving the new optimization problem. Getting rid of terms that are independent of $u$, we are actually minimizing the following functional
\begin{equation}\label{eqn:inner}
u^{(k)}=\arg\min_u ||f-u||^2+\int p_\lambda'(|\nabla u^{(k-1)}|)|\nabla u|, k=1,2,\ldots,K,
\end{equation}
which is in the same form as the functional with SATV penalty (\ref{eqn:satv}) with a weight function $w=p_\lambda'(|\nabla u^{(k-1)}|)$ that is different for each iteration $k$. Thus the computation involved is almost identical to SATV, with an extra outer loop that modifies the weight function in each iteration. Formally, each inner loop will use the evolutionary PDE derived from the Euler-Lagrange equation to solve (\ref{eqn:inner}):
\[u_t=\nabla\cdot\left\{ (p_\lambda'(|\nabla u^{(k-1)}|)\frac{\nabla u}{|\nabla u|}\right\}-(u-f).\]
 From this analogy with SATV, we can also see the advantage of SCAD from another point of view: the weight function $w=p_\lambda'(|\nabla u^{(k-1)}|)$ is bounded and thus there is no stability problem as when $w$ is inversely proportional to the first derivative, which makes an extra tuning parameter $e$ unnecessary in the SCAD model.

From the general property of the MM algorithm \cite{hunter04,nowak06}, the algorithm produces a sequence of monotonically decreasing values of the objective functional (\ref{eqn:scad}) which makes the algorithm very stable. In practice for our experiments, we find that the number of iterations $K$ can be taken as small as $K=2$, thus the running time of the algorithm is comparable to both standard TV and SATV.     

\section{Monte-Carlo SURE for Regularization Parameter Selection}
In all the above methods the value of the regularization
parameter chosen largely determines the quality of the denoised image. We use MSE as the criterion for judging the relative merits of different methods in this paper, which is defined by
\[\frac{1}{N}||u-\hat{u}||^2,\]
where we take the original image $u$ as a $N$-dimensional vector and $\hat{u}$ is the restored image. Note that it is necessary to consider discrete formulation in this section. To calculate MSE we need to have
the prior knowledge of the noise-free image which in most realistic scenarios is
unavailable. When the noise is Gaussian, \cite{ramani08} proposed a technique called Monte-Carlo SURE, which does not require any prior knowledge
of the noise-free image or the nature of the denoising algorithm. For the purpose of presenting this method, we now should change to a discrete formulation. For a noisy image $f=u+n$, formulated in the discrete domain, and a denoising algorithm considered abstractly as a mapping $\hat{u}=M(f)$ that returns a restored image $\hat{u}$ with $f$ as input, \cite{ramani08} proved that 
\begin{equation}\label{eqn:sure}
\frac{1}{N}||f-M(f)||^2-\sigma^2+\frac{2\sigma^2}{N} div_f M(f)
\end{equation}
is an unbiased estimator of the true MSE, where $\sigma$ is the standard deviation of the Gaussian noise and $div_fM(f)$ is the divergence of the multivariate function $M$. Note in our context the mapping $M$ implicitly depends on the regularization parameter $\lambda$. Direct calculation of $div_fM(f)$ is not feasible except for simple linear filtering operation, and \cite{ramani08} used the Monte Carlo approximation
\[div_fM(f)\approx\mathbf{b}^T(M(f+\epsilon\mathbf{b})-M(f)),\]
where $\mathbf{b}$ is a N-dimensional vector with i.i.d. standard normal random components, and $\epsilon$ is a small positive constant. That is, we artificially add more noise to the observed image and run the same denoising algorithm again and then approximate the divergence based on the differences of the two recovered images. We will use Monte-Carlo SURE to choose the regularization parameter whenever required in the next section. Since the noise level is assumed to be unknown in our experiments, some pilot estimate of $\sigma$ should be plugged into equation (\ref{eqn:sure}). In all our experiments, we used the following simple estimate that is quite robust empirically for blocky images:
\begin{equation}\label{eqn:sigma}
\hat{\sigma}=median\{|f_i-f_j|\}/0.954,
\end{equation}
where $f=(f_1,\ldots,f_N)$ is the observed image and the differences $f_i-f_j$ are taken over all neighboring pixels (four neighbors for each pixel). This estimate is based on the fact that with a normal random variable $X\sim N(0,2\sigma^2)$, $median(|X|)\approx 0.954\sigma$.

\section{Experiments}
\begin{figure}
\centerline{
\subfigure[]{\includegraphics[width=2.5in]{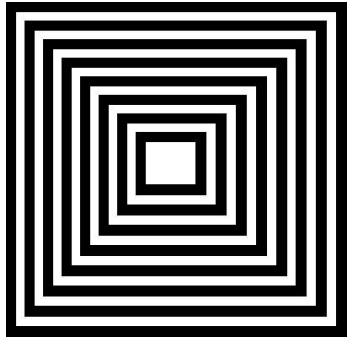}
}
}
\centerline{\subfigure[]{\includegraphics[width=2.5in]{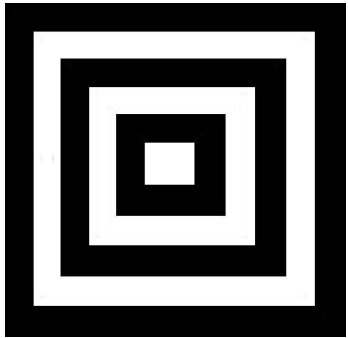}
}
\hfil
\subfigure[]{\includegraphics[width=2.7in]{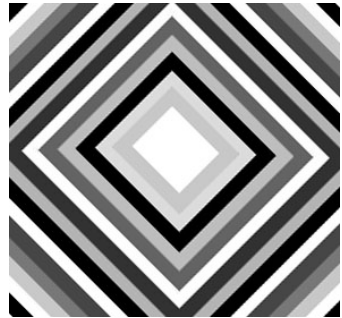}
}}
\caption{Several simple grayscale images used in the experiments. \label{box}   }
\end{figure}

\begin{figure}
\centerline{
\subfigure[]{\includegraphics[width=2.7in]{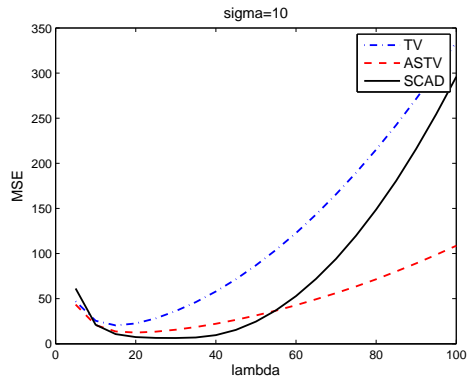}
}
}
\centerline{\subfigure[]{\includegraphics[width=2.7in]{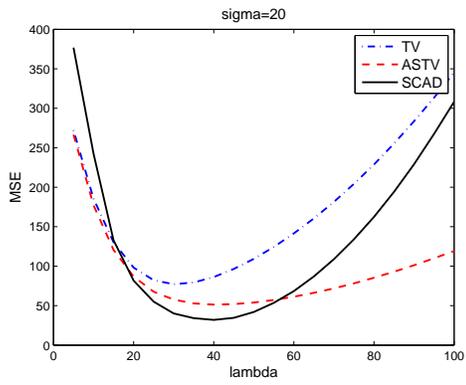}
}
\hfil
\subfigure[]{\includegraphics[width=2.7in]{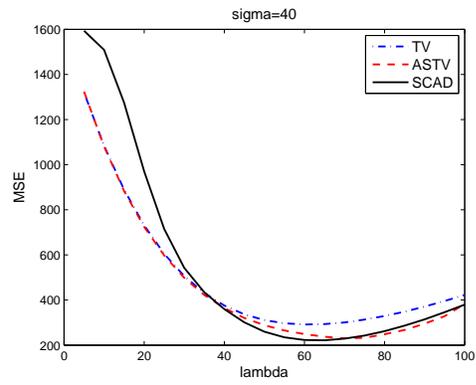}
}}
\caption{Comparison of MSE for the three methods for the image shown in Fig \ref{box}(a), with different noise levels: (a) $\sigma=10$; (b) $\sigma=20$; (c) $\sigma=40$.\label{mse}   }
\end{figure}

\begin{figure}
\centerline{
\subfigure[]{\includegraphics[width=2.7in]{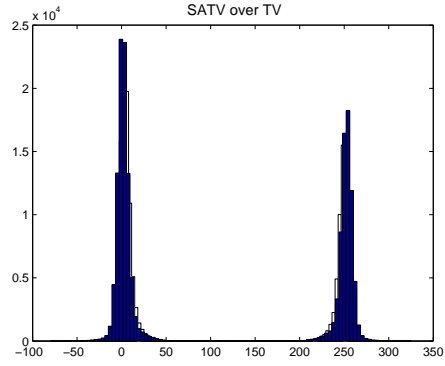}
}
}
\centerline{\subfigure[]{\includegraphics[width=2.7in]{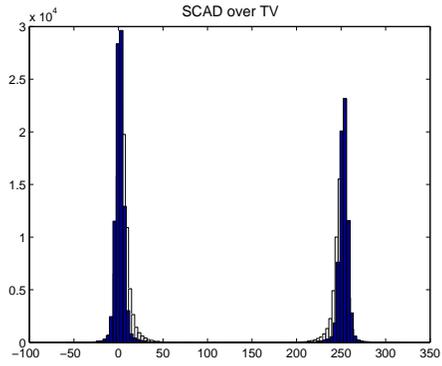}
}
\hfil
\subfigure[]{\includegraphics[width=2.7in]{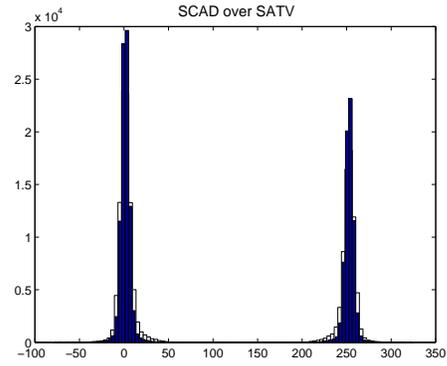}
}}
\caption{The histogram of restored image intensities overlaid on top of each other. (a) Histogram of restored image intensities obtained by SATV over that obtained by TV model. (b) Histogram of restored image intensities obtained by SCAD over that obtained by TV model. (c) Histogram of restored image intensities obtained by SCAD over that obtained by SATV model. \label{hist}   }
\end{figure}

\begin{figure}
\center{\includegraphics[width=5in]{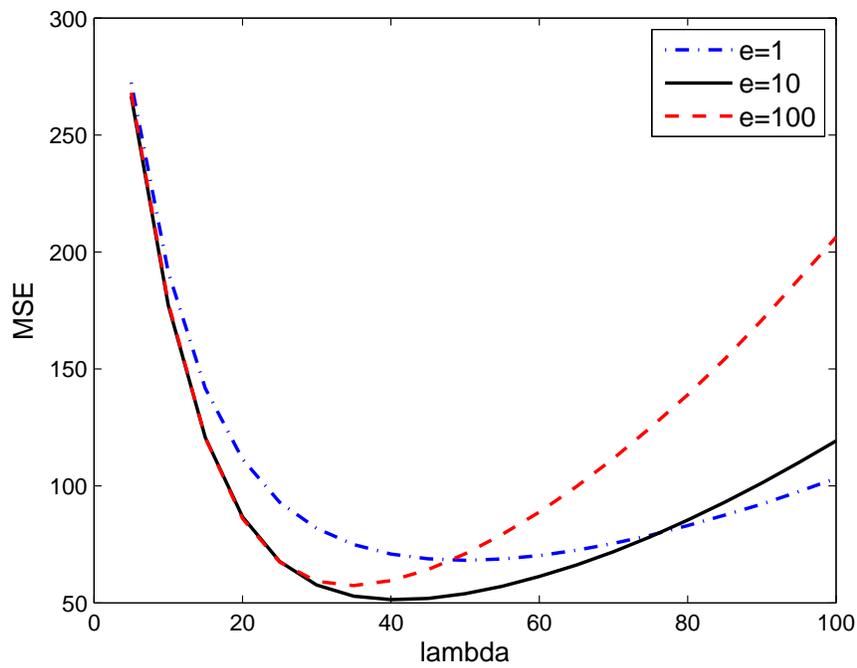}
}
\caption{Comparison of MSE for the SATV model when different values for $e$ are chosen, with noise level $\sigma=20$.\label{e}   }
\end{figure}

First we compare the performance of the three approaches TV, SATV, and SCAD using a simple black-and-white image shown in Fig \ref{box}(a). In this first experiment, we do not choose any single regularization parameter but compare the performance over a whole wide range of regularization parameters. Independent Gaussian noise with standard deviations $\sigma=10,20 \mbox{   and } 40$ are added to the original image and taken as the observed noisy input. For the initial step of SATV, we use TV with optimal parameter $\lambda$ to estimate the weight function. We also search for a good value of $e$ in the second step (based on minimization of the true MSE) for $e\in\{1,10,100,500\}$, it turns out for all three different noise levels for this image $e=10$ gives the best result. Note that we consider the intensity values of an image to be in the range of $[0,255]$. Both choices actually make the results more favorable for SATV, but we will see that even so it is being outperformed by SCAD. Fig \ref{mse} shows the evolution of the true MSE using different regularization parameters for the three methods, with different subfigures illustrating the observed image with different noise levels. From these figures, it is clearly seen that SCAD performs better than SATV, while both are significantly better than TV. To get some insights into the effect of the different penalties, the image histograms for the recovered images are shown in Fig \ref{hist} for the case of $\sigma=20$. One can see from the histograms of the TV-based restoration that the TV estimate is biased, in that black colored pixel intensities (with original intensity value of zero) are generally shifted up while white colored pixel intensities (with original intensity  of 255) are shifted down, consistent with the proposition stated previously. While SATV only partially addresses this, SCAD seems to be more efficient in solving this bias problem.  Besides, Fig \ref{hist}(c) demonstrates that for the recovered image using the SCAD penalty, the histogram is more peaked and thus resulting in smaller MSE.  
Using this experiment, we can also see the effect of $e$ on the result. As stated above $e=10$ is optimal for SATV for this image. We see from Fig \ref{e} that using $e=1$ or $e=100$ makes the MSE bigger. Specifically, using $e=1$ enlarged the minimum MSE from 51.40 to 68.21, or by $34\%$, while using $e=100$ enlarged MSE by $13\%$. Unfortunately there is no universally best value for $e$, and our later experiments demonstrate that for different images the optimal $e$ is difficult to predict. Choosing a wrong value for $e$ makes the performance of SATV more unpredictable. Although $e$ could be selected by similar methods that have been developed for selecting $\lambda$, for example using Monte-Carlo SURE, this at least increases significantly the computational burden of the algorithm. And even with a good estimate of $e$, our result here shows that it is still worse than SCAD in terms of the MSE criterion.

\begin{figure}
\center{\includegraphics[width=5in]{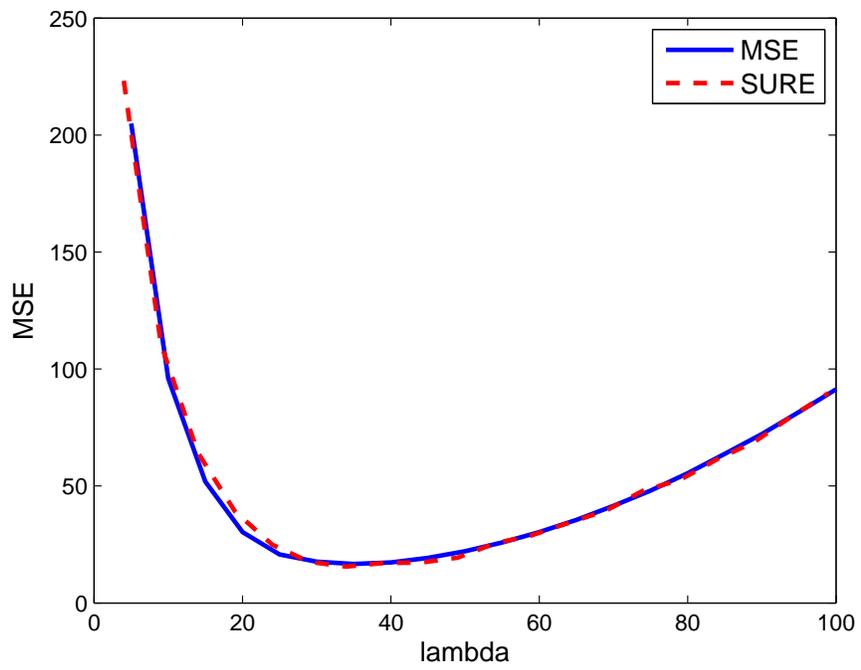}
}
\caption{MSE and SURE estimate for the SCAD method.\label{sure}   }
\end{figure}

Our second experiment uses images as shown in Fig \ref{box} (b) and (c). The former is still a black-and-white image with thicker nested squares. The latter is an image similar in structure to Fig \ref{box}(a) but with different grayscale levels and also rotated by $45^o$ degrees. Image Fig \ref{box}(b) is clearly easier to denoise due to the larger scale of its features, thus we choose to add Gaussian noise with standard deviations $\sigma=20,40,80$. For image (c) we use four different levels $\sigma=10,20,40,80$. The regularization parameters now are selected using Monte-Carlo SURE as briefly described previously with $\sigma$ assumed unknown and estimated using (\ref{eqn:sigma}). The effectiveness of Monte-Carlo SURE in general has been demonstrated for some methods including TV model in \cite{ramani08}. We additionally verified its performance in our SCAD model under several situations and found it to be quite accurate for our proposed model. As an illustration, for denoising the image shown in Fig \ref{box}(b) with $\sigma=20$, we demonstrate that Monte-Carlo SURE accurately predicts the true MSE in Fig \ref{sure}. The MSE of the restoration results for the two images are shown in Table \ref{tab:b} and \ref{tab:c} respectively. For the SATV method, the optimal values of $e$ in each situation is also indicated in the table. Note that the optimal $e$ is found from the true MSE and thus the results presented is favorable for the SATV method. The reader can now see that different situations require different choices of $e$ and there seems to be no universal way of specifying a good value a priori. The conclusion is the same as before: SCAD is superior to SATV.

\begin{figure}
\centerline{
\subfigure[]{\includegraphics[width=2.7in]{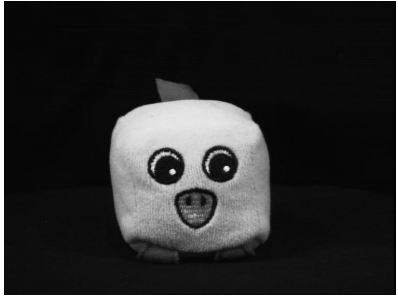}
}
\hfil
\subfigure[]{\includegraphics[width=2.7in]{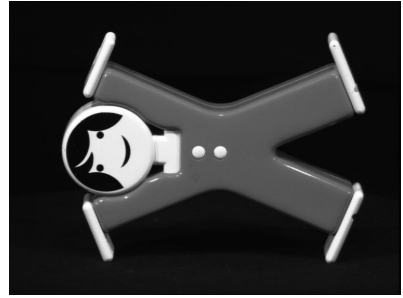}
}}
\centerline{\subfigure[]{\includegraphics[width=2.7in]{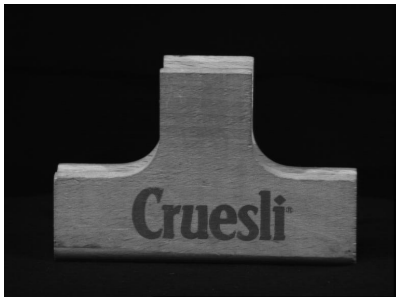}
}
\hfil
\subfigure[]{\includegraphics[width=2.7in]{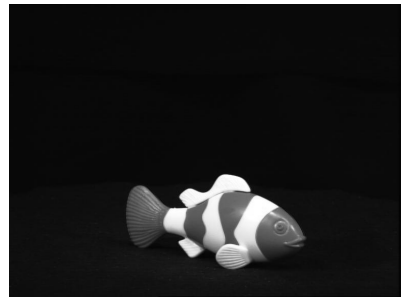}
}}
\caption{Four images obtained from ALOI used for testing the performances of different methods.\label{obj}   }
\end{figure}

Finally, we use some slightly more complicated images to test the performances. Amsterdam Library of Object Images (ALOI, \url{ http://staff.science.uva.nl/~aloi/}) is a color image collection of one-thousand small objects, recorded for scientific purposes. We pick four images as shown in Fig \ref{obj} and transform them to grayscale images, which looks close to piece-wise constant visually. Gaussian noises with standard deviation of $40$ are added to each image and different methods are applied. The results in terms of MSE are shown in Table \ref{tab:object}, and the method using the SCAD penalty is still the best even for these more complicated images. Since it is visually difficult to distinguish the restored images in print using different methods, we choose not to show the restored images here, but the images are available from \url{http://?} in MATLAB's .fig format. 

\section{Conclusion}
In this paper, we proposed a new penalization functional for image denoising. The penalty function is directly motivated by the well-known oracle property of the SCAD penalty from the statistical literature originally proposed for high-dimensional statistical regression problems. Using a simple argument in a maybe overly simplistic situation, i.e., our two-pixel image model (\ref{eqn:twopixel}), we show that the  functional with SCAD penalty solves the bias problem inherent in TV regularization, which is also verified by our experimental results. Compared to spatially adaptive TV, the newly proposed method gets rid of the headache of choosing an extra parameter that controls the stability and adaptivity of the algorithm, and achieves better mean squared error at the same time. Our goal in this paper is not to propose a general image denoising method to compete with the state-of-the-art such as the wavelet-based method or the nonlocal mean \cite{buades05} which has become very popular recently, but to show that a carefully designed penalty function can improve existing PDE-based approaches without extra computational burden. Due to its shrinkage to zero of the first order differences, the method is most suitable for recovering blocky images. One can also penalize higher order derivatives as has been done for TV regularization, but this is outside the scope of the current paper.

\section*{Appendix}
We only prove the proposition for parts (a) and (b), the proofs for parts (c) and (d) are similar and slightly simpler. 
Let $Q(\theta_1,\theta_2)=(y_1-\theta_1)^2+(y_2-\theta_2)^2+p_\lambda(|\theta_1-\theta_2|)$. Obviously the minimizer satisfies $\theta_1\ge\theta_2$ when $y_1\ge y_2$ (otherwise exchanging the values of $\theta_1$ and $\theta_2$ makes the functional smaller).
The partial derivatives are (for $\theta_1>\theta_2$)
\begin{eqnarray*}
\frac{\partial Q}{\partial\theta_1}&=&2(\theta_1-y_1)+p_\lambda'(|\theta_1-\theta_2|),\\
\frac{\partial Q}{\partial\theta_2}&=&2(\theta_2-y_2)-p_\lambda'(|\theta_1-\theta_2|).\\
\end{eqnarray*}
The complication only comes from nondifferentiability when $\theta_1=\theta_2$. When constrained to $\theta_1=\theta_2$, it is easy to see from the quadratic form of $Q$ that the only potential minimizer is $\theta_1=\theta_2=(y_1+y_2)/2$. Meanwhile, when $y_1-y_2>a\lambda$, we have $Q((y_1+y_2)/2,(y_1+y_2)/2)=(y_1-y_2)^2/2>(a+1)\lambda^2/2=p_\lambda(|y_1-y_2|)=Q(y_1,y_2)$. Thus the minimizer must satisfy $\theta_1\neq\theta_2$ and the functional is differentiable near the minimizer, which in turns implies that both partial derivatives are equal to zero. 
Adding and subtracting the two partial derivatives, we get 
\begin{eqnarray}
\theta_1+\theta_2&=&y_1+y_2,\label{eqn:add}\\
\theta_1-\theta_2&=&y_1-y_2-p_\lambda'(|\theta_1-\theta_2|).\label{eqn:subtract}
\end{eqnarray}
From (\ref{eqn:subtract}), $\theta_1-\theta_2$ is a solution to the equation $x+p_\lambda'(x)=y_1-y_2$. The function on the left hand side, when written down explicitly, is
\begin{equation}\label{eqn:lhs}
x+p_\lambda'(x)=\left\{\begin{array}{cc}
\lambda+x & x<\lambda\\
\frac{a\lambda}{a-1}+(1-\frac{1}{a-1})x& \lambda\le x\le a\lambda\\
x& x>a\lambda
\end{array}\right.
\end{equation}
which is strictly increasing for $x>0$ and the equation $x+p_\lambda'(x)=y_1-y_2$ obviously has a unique solution $x=y_1-y_2$ when $y_1-y_2>a\lambda$. Combine this with (\ref{eqn:add}), we get $\theta_1=y_1,\theta_2=y_2$, and part (a) is proved.

For part (b), if the minimizer satisfies $\theta_1\neq\theta_2$ so that the minimizer is a stationary point, then $\theta_1-\theta_2> 0$ is a solution to the equation $x+p_\lambda'(x)=y_1-y_2$ by exactly the same arguments as before. From (\ref{eqn:lhs}), it is easy to see that the left hand side is bounded below by $\lambda>0$ and thus there exists no solution when $y_1-y_2<\lambda$, leading to a contradiction. Now with the constraint $\theta_1=\theta_2$, it is immediate from the form of the functional $Q(\theta_1,\theta_2)$ that $\theta_1=\theta_2=(y_1+y_2)/2$.

\bibliographystyle{elsarticle-num}
\bibliography{papers.txt,books.txt}

\newpage

\begin{table}
\center
\begin{tabular}{cccc}
noise level & TV &ASTV&SCAD\\
$\sigma=20$ &31.97&24.96(e=100)&17.13\\
$\sigma=40$ &114.71&95.76(e=100)&92.00\\
$\sigma=80$ &415.11&387.10(e=100)&383.77\\
\end{tabular}
\caption{MSE of using different methods on the image shown in Fig \ref{box}(b).\label{tab:b}}
\end{table}

\begin{table}
\center
\begin{tabular}{cccc}
noise level & TV &ASTV&SCAD\\
$\sigma=10$ &37.02&34.10 (e=10)&29.10\\
$\sigma=20$ &99.37&92.46(e=10)&77.39\\
$\sigma=40$ &370.68&275.08(e=10)&266.65\\
$\sigma=80$ &886.95&858.32(e=100)&805.66\\
\end{tabular}
\caption{MSE of using different methods on the image shown in Fig \ref{box}(c).\label{tab:c}}
\end{table}

\begin{table}
\center
\begin{tabular}{cccc}
        & TV &ASTV&SCAD\\
$duck$ &77.20&75.80 (e=100)&69.70\\
$person$ &93.22&84.89(e=100)&79.35\\
$board$ &82.58&74.95(e=100)&68.87\\
$fish$ &70.99&63.69(e=100)&55.58\\
\end{tabular}
\caption{The MSE for different methods applied to four object images obtained from ALOI when $\sigma=40$.\label{tab:object}}
\end{table}

%
%
%
\end{document}